\documentclass{article}
\usepackage{spconf,amsmath,amssymb,graphicx,multirow}
\usepackage{balance}

\usepackage{enumitem}
\setlist{nosep, leftmargin=14pt}

\usepackage{mwe} 

\title{Reconstruction of Surface EMG Signal using IMU data for Upper Limb Actions}
\name{Shubhranil Basak$^{\star}$ \qquad Mada Hemanth$^{\star}$ \qquad Madhav Rao$^{}$\thanks{$\star$ The authors contributed equally to this research work.}}

\address{International Institute of Information Technology-Bangalore}

\begin{document}

\maketitle

\begin{abstract}
Surface Electromyography (sEMG) provides vital insights into muscle function, but it can be noisy and challenging to acquire. Inertial Measurement Units (IMUs) provide a robust and wearable alternative to motion capture systems. This paper investigates the synthesis of normalized sEMG signals from 6-axis IMU data using a deep learning approach. We collected simultaneous sEMG and IMU data sampled at 1~KHz for various arm movements. A Sliding-Window-Wave-Net model, based on dilated causal convolutions, was trained to map the IMU data to the sEMG signal. The results show that the model successfully predicts the timing and general shape of muscle activations. Although peak amplitudes were often underestimated, the high temporal fidelity demonstrates the feasibility of using this method for muscle intent detection in applications such as prosthetics and rehabilitation biofeedback.
\end{abstract}


\section{Introduction}
\label{sec:intro}

Surface Electromyography (sEMG) is quickly becoming important in various applications, including biomedical/clinical settings, prosthetic and rehabilitation devices, and human-machine interactions. Electromyography (EMG) Signals are the electrical signals generated due to muscle contractions. sEMG is a non-invasive technique employed to read EMG signals. Although effective, the data can be affected by interference (noise) as the signal must pass through the skin. EMG Signals are very vital for assessing muscle function during movement, such as walking and sports. It also plays a crucial role in physical therapy by providing biofeedback to help patients re-educate specific muscles. Normalized sEMG can be used to compare muscle activation patterns between individuals over time with high reliability, but does not allow comparisons of activity levels between muscles, tasks, or individuals~\cite{normalization_emg}. An Inertial Measurement Unit (IMU) is a device used to capture the movement of a body using a combination of an Accelerometer, a Gyroscope, and a Magnetometer. In contrast to the noisy data from sEMG sensors, IMUs are robust and can be easily integrated into wearable devices. This convenience led us to investigate the possibility of reconstructing the sEMG signal using the IMU signal. If an accurate mapping between the IMU data and the sEMG signal is possible, it would enable the development of muscle activity monitoring without the need for moderately expensive sEMG instrumentation. 
Additionally, a customized sEMG wearable that fits all individual physicalities is not feasible; hence, IMU-driven sEMG signal reconstruction is a suitable alternative.
Advances made in Machine Learning, particularly deep learning for time-series analysis, have shown great progress in mapping complex, non-linear relationships. We have utilized WaveNet, a deep neural network architecture developed by Google DeepMind, primarily used for generating audio signals. Our proposed model utilizes a sliding window to learn the mapping from 6-axis IMU data to the normalized sEMG signal. The primary contribution of this work is the evaluation and development of a deep learning model for the synthesis of sEMG signal from IMU data.

\section{Data Acquisition}
\label{sec:acquisition}

Surface EMG data for this research were collected from 2 healthy subjects 20 years old. A self-declaration with no prior history of health issues and surgeries on any parts of their body was also collected from the two subjects.
The study
was ethically approved by the Institutional Ethics Committee
(IEC). Informed ethical consent was taken from the
healthy subject with a declaration approving that the subject
had no prior weakness or injury. Data was collected
in accordance with the Helsinki Declaration of 1975, as
revised in 2000~\cite{helsinki}. 
For collecting sEMG data, we used a muscle Bioamp patch with a sampling frequency of 1 kHz. All motions were performed in a certain number of sets. For data collection from biceps curl and triceps extension, the motions were performed in a set of 7 contractions and relaxation movements. For supination and pronation, they were performed until the subject experienced fatigue in the target muscle. Data collection for each motion was conducted over a period of 4 days to capture intra-subject variability. In addition to the sEMG data, we utilized an MPU 6050, a 3-axis gyroscope and accelerometer sensor, to capture motion data. All data were stored in a columnar manner as text and CSV files.

\begin{figure}[h]
    \centering
    \includegraphics[width=1\linewidth]{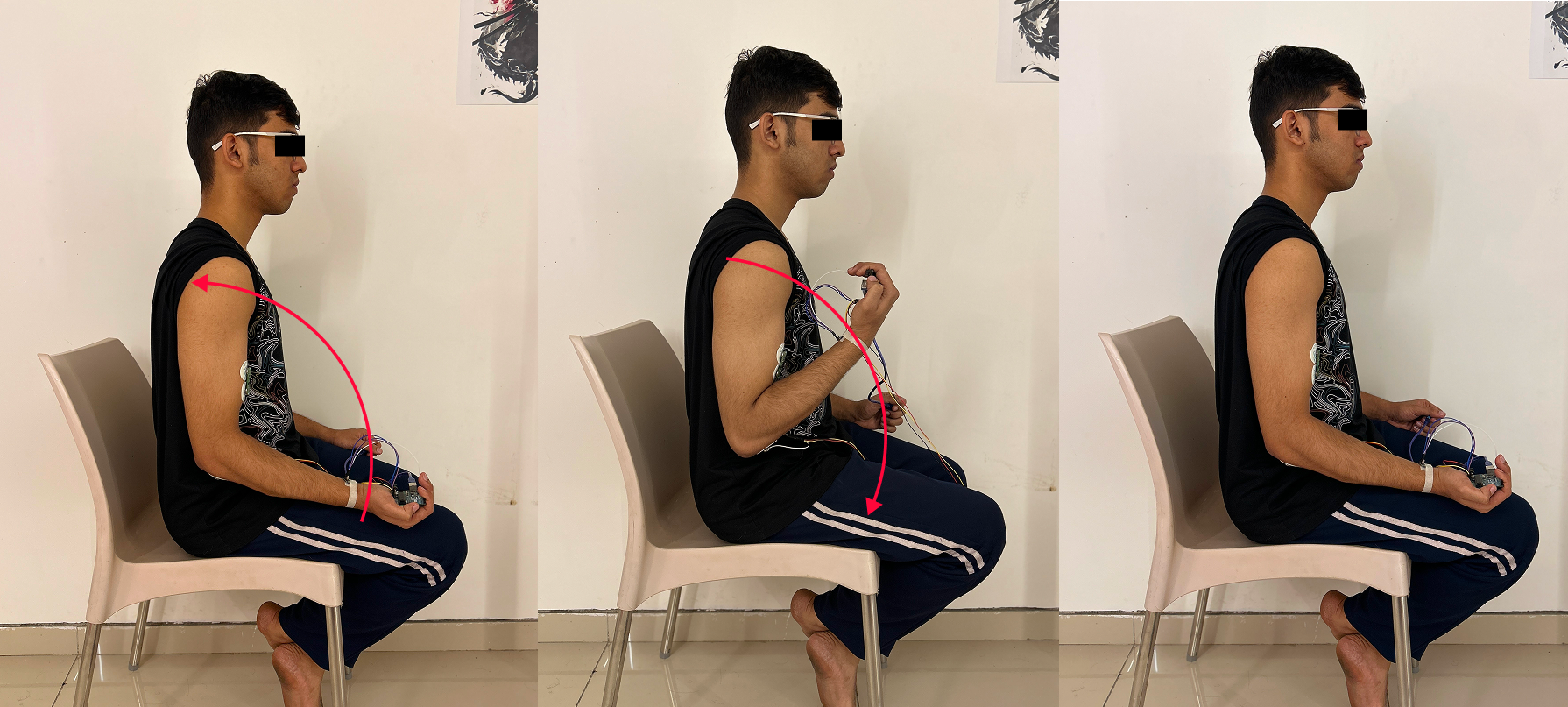}
    \caption{Snapshot of an individual performing Bicep Curl Exercise.}
    \label{fig:Bicep_Curl_exercise}
\end{figure}

Fig~\ref{fig:Bicep_Curl_exercise} demonstrates the movement and setup \cite{IMU_Placement} adopted for the bicep curl exercise. Similarly, tricep extension, supination, and pronation exercises were also conducted. The positioning of electrodes and movement \cite{estimation_resistance} to obtain sEMG from the target muscle for each of these upper limb motions were carried out according to the research conducted by SENIAM~\cite{seniam}.

\section{Data Pre-processing and Segmentation}
\label{sec:pagestyle}
\subsection{Filtering}
\label{ssec:filtering}
After collecting the IMU and sEMG data, pre-processing was performed using a series of digital 4-order Butterworth filters to remove noise and extract valid muscle activity features.

\begin{enumerate}
    \item \textbf{High-pass filter:} A 70~Hz high-pass filter was used to eliminate motion artifact commonly arising from electrode movement. \cite{filters}

    \item \textbf{Band-pass filter:} A range of 20~Hz to 300~Hz was set to preserve physiologically relevant EMG frequency components.
     \cite{filters}

    \item \textbf{Notch filter:} A narrow band-stop filter between 48-52~Hz was used to suppress power line interference.
     \cite{filters}
\end{enumerate}

\vspace{-5mm}

\subsection{Segmentation}
\label{ssec:segmentaion}
After filtering, we performed peak-based segmentation. Positive peaks were identified with a minimum peak distance of 150 samples\cite{segmentation}. The seven highest peaks were detected and boundaries were defined at the midpoints between successive peaks. This enabled the automatic isolation of muscle movement segments. The IMU data was also segmented using the same index ranges.

\vspace{-5mm}

\subsection{Data merge}
\label{ssec:data merge}
Each segment was then enriched with simultaneously recorded IMU features. Accelerometer and IMU data from all three axis (X, Y, and Z) were aligned with EMG indices and appended to each segment entry. After adding metadata, all processed segments were consolidated into a unified CSV dataset for downstream model training. 

\begin{figure}[h!]
    \centering
    \includegraphics[width=1\linewidth]{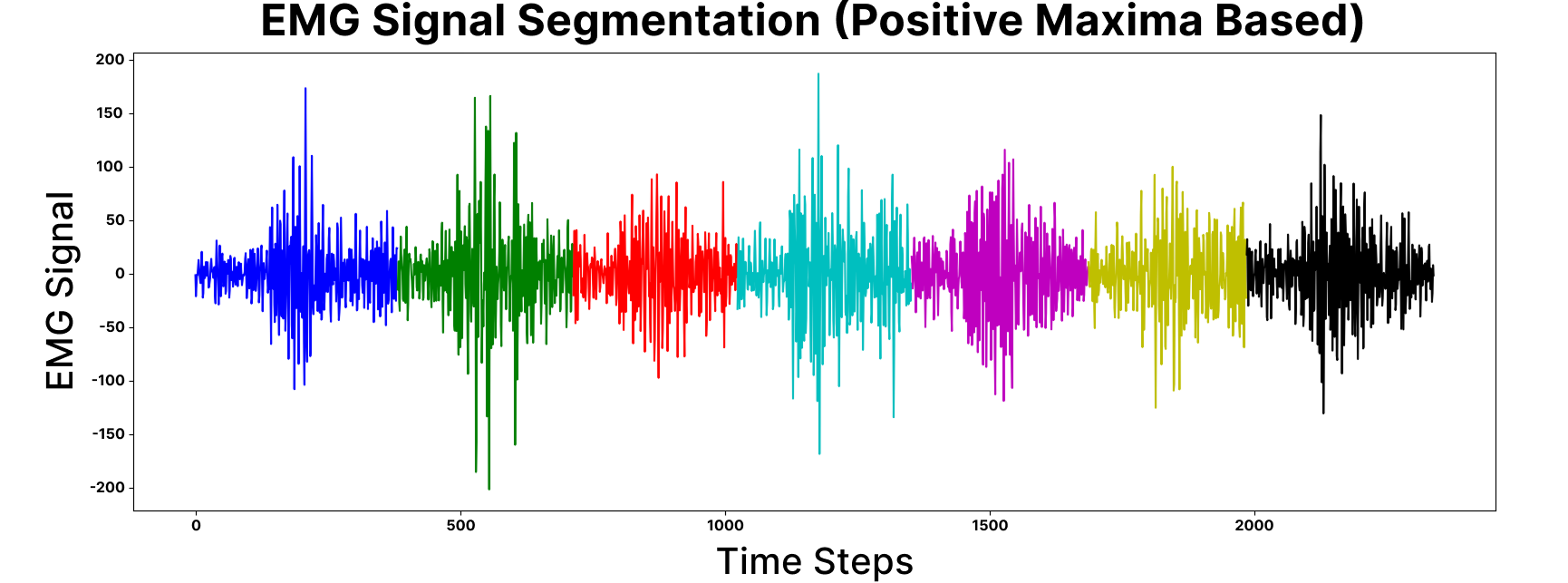}
    \caption{ Segmented Plot for seven sets of bicep curl}
\end{figure}

\section{Proposed Model Architecture}
\label{sec:model}

We propose a sliding window WaveNet\cite{wavenet} architecture with stacked WaveNetblocks inspired from the Transformer\cite{AIAYN} architecture that maps IMU motion signals to EMG muscle activation patterns\cite{EMG_wrist}. The network leverages dilated causal convolutions\cite{cnn_dilation} to capture both local kinematic responses and long-range temporal dependencies while maintaining strict causality\cite{cnn_dilation}, a critical requirement for real-time biomechanical applications.

\subsection{Network Overview}

The architecture consists of four main components: (1) an input projection layer that transforms multi-dimensional IMU features into a unified latent representation, (2) a stack of dilated causal convolution blocks\cite{dilation} that progressively expand the receptive field, (3) a context aggregation module with sliding window convolution, and (4) an output projection layer that generates single-channel EMG predictions. The network processes variable-length sequences through a fully convolutional design, eliminating the need for recurrent connections and enabling efficient parallel computation.

\subsection{Input Layer}

The input layer performs a $1\times1$ convolution on raw IMU features to project them into a higher-dimensional latent space. This learned linear projection combines information across different sensor modalities without mixing temporal information, allowing the network to discover optimal feature combinations before temporal processing.

\subsection{Dilated Causal Convolution Blocks}

The core architecture consists of $N$ stacked WaveNet blocks, each of which employees dilated causal convolutions\cite{dilation} with exponentially increasing dilation factors ($d_i = 2^i$). This enables exponential growth of the receptive field with linear increase in network depth, while each stacked block adds wider context to the outputs of the previous Wavenet Block\cite{cnn_dilation}.

\begin{figure*}[t]
    \centering
    \includegraphics[width=\textwidth]{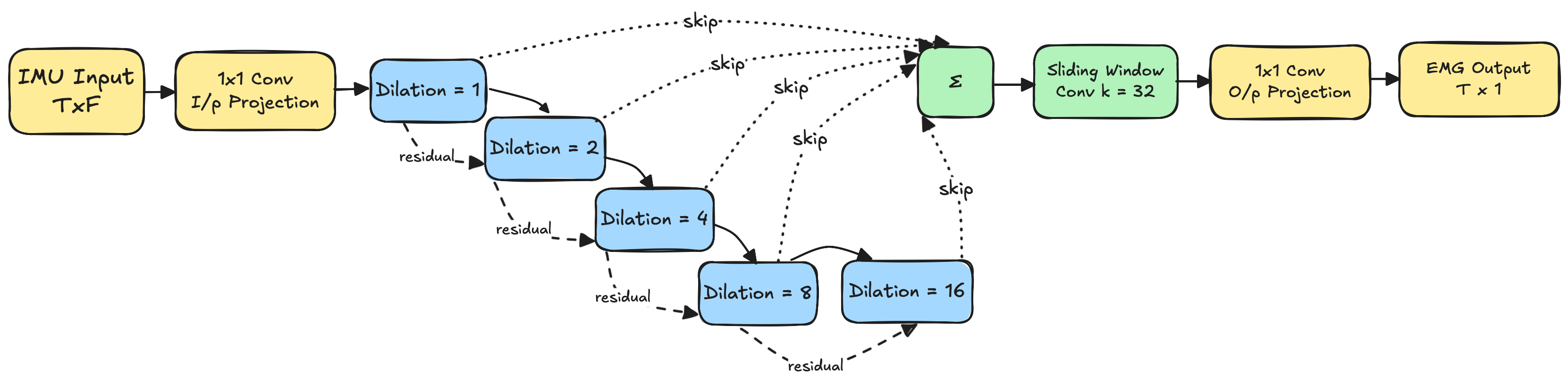}
    \caption{Model Block Diagram}
    \label{fig:model}
\end{figure*}

\begin{figure*}[t]
    \centering
    \includegraphics[width=\textwidth]{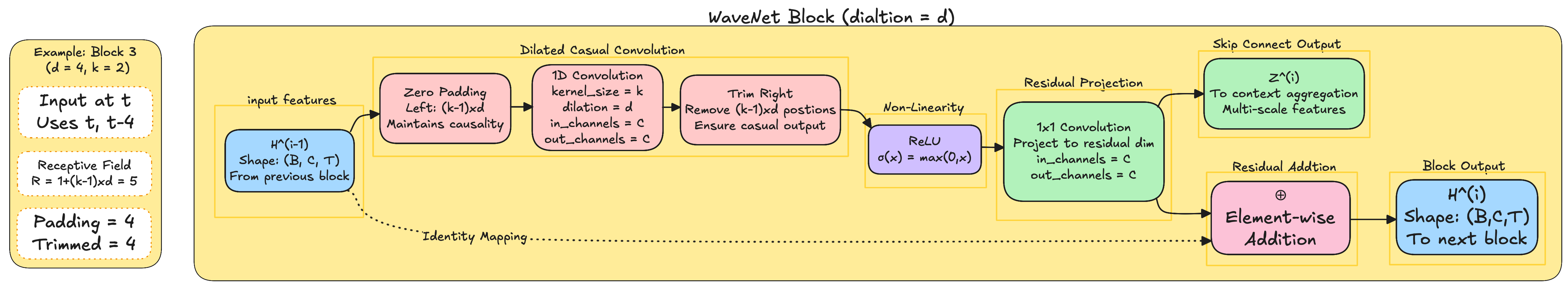}
    \caption{Dilation Block}
    \label{fig:dilation}
\end{figure*}

Each WaveNetBlock applies:
\begin{itemize}
    \item \textbf{Dilated Causal Convolution}: Ensures predictions at time $t$ depend only on inputs from $t' \leq t$. Padding of $(k-1) \cdot d_i$ maintains causality.
    \item \textbf{Residual Connection}: A $1\times1$ convolution projects activated features back to the residual dimension before addition, facilitating gradient flow.
\end{itemize}

The receptive field after $N$ blocks is:
\begin{equation}
R = 1 + (k-1)(2^N - 1)
\end{equation}

\subsection{Context Aggregation Layer}

A key architectural contribution is the context aggregation module that combines multi-scale temporal features. Skip connections from all dilated blocks are summed:

\begin{equation}
\mathbf{S} = \sum_{i=1}^{N} \mathbf{H}^{(i)}
\end{equation}

These aggregated features are processed through a sliding window convolution with window size $w$:

\begin{equation}
\mathbf{C} = \text{Conv}_w(\mathbf{S})
\end{equation}

This layer integrates information across a fixed temporal neighborhood, providing robustness to local variations in movement dynamics—particularly important for capturing the biomechanical relationship between IMU signals and muscle activation.

\subsection{Output Layer}

A final $1\times1$ convolution projects contextualized features to single-channel EMG predictions without activation functions, allowing unbounded amplitude predictions. The model is trained end-to-end using MSE loss.

\section{Results}
\label{sec:results}
\subsection{Qualitative Analysis}
\begin{figure*}[t]
    \centering
    \includegraphics[width=1\textwidth]{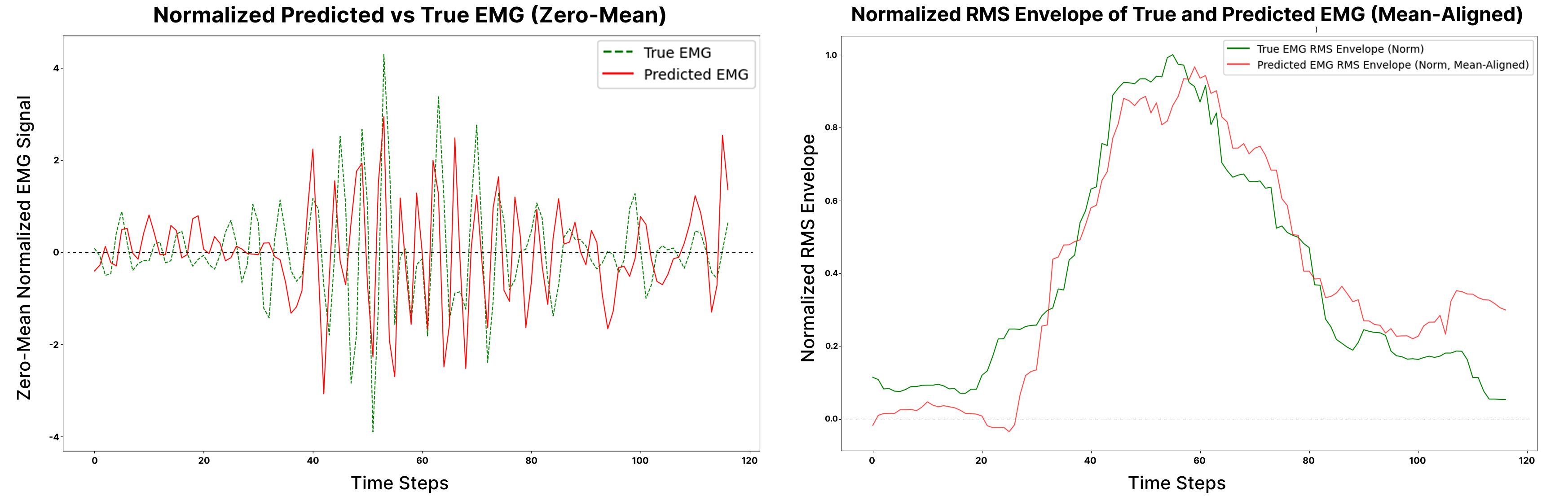}
    \caption{Synthesis of Normalized sEMG signal (for Bicep Curl) and its corresponding envelope waveform.}
    \label{fig:sample_prediction_image}
\end{figure*}

A Qualitative analysis between predicted and true sEMG shown in Fig~\ref{fig:sample_prediction_image} reveals a high temporal fidelity. The model was able to learn the temporal relationship between IMU data and sEMG data and accurately predict the burst timing. The predicted signal burst and the corresponding relaxation phase align well with the true sEMG envelope. Although temporal characteristics and the overall shape are well captured, the model struggles to achieve the same peak amplitude as the true sEMG, as seen in Fig~\ref{fig:sample_prediction_image}, where the predicted peak is noticeably lower than the actual signal's peak.

\subsection{Quantitative Analysis}
Multiple models were trained, each targeting a specific muscle movement. All the models were trained on 85\% of the dataset and was tested on the remaining 15\% of the dataset with an early stopping patience of 5, the model was trained for 119 epochs. Table ~\ref{tab:emg_results} reports the results we have obtained.

\subsection{Discussion of Model Performance}
The prediction of the normalized sEMG envelope from IMU data demonstrated a critical dichotomy in model fidelity. Quantitative analysis\cite{metric} across four muscle groups revealed consistently high \textbf{Time-Domain Cosine Similarity} ($\rho_{t} > 0.91$), confirming that the IMU successfully captures the accurate timing and temporal template of muscle activation. However, this temporal success was undermined by a systematic failure in dynamic synthesis, evidenced by a dramatic reduction in the \textbf{FFT Cosine Similarity} (e.g., $\rho_{f}$ dropping to $0.4245$ for the Triceps). This spectral deficiency indicates that the model produces an overly smoothed, low-pass filtered output that lacks the high-frequency dynamics essential for precise force grading. Magnitude accuracy, measured by the \textbf{Mean Absolute Error (MAE)}, was competitive (Bicep MAE $= 0.1643$) but exhibited significant functional heterogeneity: performance was highest for the simple flexor (Bicep) and lowest in bio-mechanically complex actuators (Triceps and rotational groups), which also showed low robustness and high variability. In conclusion, the model provides a reliable signal for movement \textit{timing} and \textit{classification}, but its current architecture is insufficient for applications requiring high-fidelity proportional control due to poor spectral replication.

\begin{table}[h]
\centering
\caption{Surface EMG Envelope Prediction Results Across Upper Limb Motions from respective Target Muscles}
\begin{tabular}{|p{1.5cm}|p{1.75cm}||p{0.85cm}|p{1cm}|p{1.2cm}|}
\hline
\multicolumn{2}{|c||}{\textbf{Motion}} & \multicolumn{3}{c|}{\textbf{EMG Envelope Metrics}} \\
\hline
\textbf{Motion} & \textbf{Metric} & \textbf{Best} & \textbf{Worst} & \textbf{Average} \\
\hline
\multirow{4}{2.5cm}{Bicep curl} 
    & MSE & 0.0237 & 0.0738 & 0.0459 \\
    & MAE & 0.1097 & 0.2215 & 0.1643 \\
    & Cosine Sim & 0.9574 & 0.8990 & 0.9280 \\
    & FFT Cosine & 0.8617 & 0.5193 & 0.7206 \\
\hline
\multirow{4}{2.5cm}{Tricep \\flection} 
    & MSE & 0.0318 & 0.1109 & 0.0592 \\
    & MAE & 0.1335 & 0.2975 & 0.1933 \\
    & Cosine Sim & 0.9637 & 0.8926 & 0.9321 \\
    & FFT Cosine & 0.6150 & 0.3312 & 0.4245 \\
\hline
\multirow{4}{2.5cm}{Supination} 
    & MSE & 0.0158 & 0.1607 & 0.0707 \\
    & MAE & 0.0892 & 0.3822 & 0.2174 \\
    & Cosine Sim & 0.9764 & 0.8831 & 0.9427 \\
    & FFT Cosine & 0.8385 & 0.5666 & 0.6909 \\
\hline
\multirow{4}{2.5cm}{Pronation} 
    & MSE & 0.0156 & 0.1627 & 0.0643 \\
    & MAE & 0.0880 & 0.3728 & 0.2101 \\
    & Cosine Sim & 0.9537 & 0.8482 & 0.9160 \\
    & FFT Cosine & 0.8673 & 0.6533 & 0.7414 \\
\hline
\end{tabular}

\label{tab:emg_results}
\end{table}

\section{Conclusion}
\label{sec:conclusion}

In conclusion, our work demonstrates the viability of predicting the normalized sEMG signal from the corresponding IMU data. The model demonstrated its effectiveness in learning the temporal relationship between motion and muscle activity, accurately predicting the timings of muscle contraction and relaxation. This exceptional temporal performance indicates that the model is well-suited for applications where detecting muscle intent is key, such as in prosthetic control or assistive devices.
The most significant limitation of the model is its inability to precisely match the peak amplitude of the accurate sEMG signal, even when the timing is correct. This implies that while the model excels at reconstructing the signal when the muscle is active, it is less accurate in predicting the strength of the signal.
Future work will focus on improving this difference in the amplitude of the predicted and actual sEMG signal. Additionally, this study could be expanded by collecting data from a larger and more diverse cohort of subjects and by testing the model's ability to generalize for diverse movements.
Finally, optimizing the model for computational efficiency would be a critical next step towards implementing this system in a real-time environment for applications such as prosthetic control or rehabilitation biofeedback. \cite{rehab}


\end{document}